\documentclass{article}

\usepackage{arxiv}
\usepackage{kotex}
\usepackage[utf8]{inputenc} % allow utf-8 input
\usepackage{multirow}
\usepackage[T1]{fontenc}    % use 8-bit T1 fonts
\usepackage{hyperref}       % hyperlinks
\usepackage{url}            % simple URL typesetting
\usepackage{booktabs}       % professional-quality tables
\usepackage{amsfonts}       % blackboard math symbols
\usepackage{nicefrac}       % compact symbols for 1/2, etc.
\usepackage{lipsum}		% Can be removed after putting your text content
\usepackage{graphicx}
\usepackage{doi}
\usepackage{algorithm}
\usepackage{algpseudocode}

\title{Fast Training of NMT Model with Data Sorting}

\date{} 					% Or removing it

\author{ Daniela N. Rim \\ 
	\textit{Handong Global University}\\
	Republic of Korea\\
        \texttt{rim.dan96@gmail.com}\\ 
	\And
        Kimera Richard \\
   \textit{Handong Global University}\\
    Republic of Korea\\
    \texttt{kimrichies@handong.ac.kr}\\
	\And
    Heeyoul Choi \\
	\textit{Handong Global University}\\
	Republic of Korea\\
	\texttt{heeyoul@gmail.com} \\
}

\renewcommand{\headeright}{}
\renewcommand{\undertitle}{}

\hypersetup{
pdftitle={A template for the arxiv style},
pdfsubject={cs.ML},
pdfauthor={Daniela Rim},
pdfkeywords={},
}

\begin{document}
\maketitle
\begin{abstract}
The Transformer model has revolutionized Natural Language Processing tasks such as Neural Machine Translation, and many efforts have been made to study the Transformer architecture, which increased its efficiency and accuracy. One potential area for improvement is to address the computation of empty tokens that the Transformer computes only to discard them later, leading to an unnecessary computational burden. To tackle this, we propose an algorithm that sorts translation sentence pairs based on their length before batching, minimizing the waste of computing power. Since the amount of sorting could violate the independent and identically distributed (i.i.d) data assumption, we sort the data partially. In experiments, we apply the proposed method to English-Korean and English-Luganda language pairs for machine translation and show that there are gains in computational time while maintaining the performance. Our method is independent of architectures, so that it can be easily integrated into any training process with flexible data lengths.  
%The Transformer model has revolutionized Natural Language Processing tasks such as Neural Machine Translation. Many efforts have been made to study the Transformer architecture, which increased its efficiency and accuracy. One potential area for improvement is to address the computation of empty tokens that the Transformer computes only to discard them later, leading to an unnecessary computational burden. To tackle this, we propose an algorithm that sorts translation sentence pairs based on their length before batching, minimizing the waste of computing power. Since the amount of sorting could violate the independent and identically distributed (i.i.d) data assumption, we propose a partial sorting algorithm and apply it to English-Korean and English-Luganda language pairs for machine translation. We empirically show that there are gains in computational time while maintaining the performance. Our method can be easily integrated into any Transformer architecture for other tasks.
\end{abstract}

% keywords can be removed

\vspace{4em}

\section{Introduction}
With the introduction of the Transformer model \cite{vaswani2017attention}, Natural Language Processing (NLP) tasks such as Neural Machine Translation (NMT) have seen exponential increases in performance. Furthermore, the Transformer architecture is at the core of the state-of-the-art models such as BERT \cite{devlin2018bert}, GPT-3 \cite{brown2020language}, mT6 \cite{chi2021mt6}, among many others \cite{kalyan2021ammus}. Due to its importance in many applications, several consecutive studies of the Transformer architecture give insight on the role of each individual component \cite{lin2022survey}. These studies result in potential improvements that could increase the efficiency and accuracy of the Transformer by changing the modules (attention mechanism, feedforward layers, etc.), architecture, pre-training, and many others \cite{han2021learning, kitaev2020reformer, sukhbaatar2019adaptive, wang2019r,sukhbaatar2019augmenting,wu2020lite, liu2021swin}.

\begin{figure}[ht]
	\centering
	\includegraphics[width=0.4\linewidth]{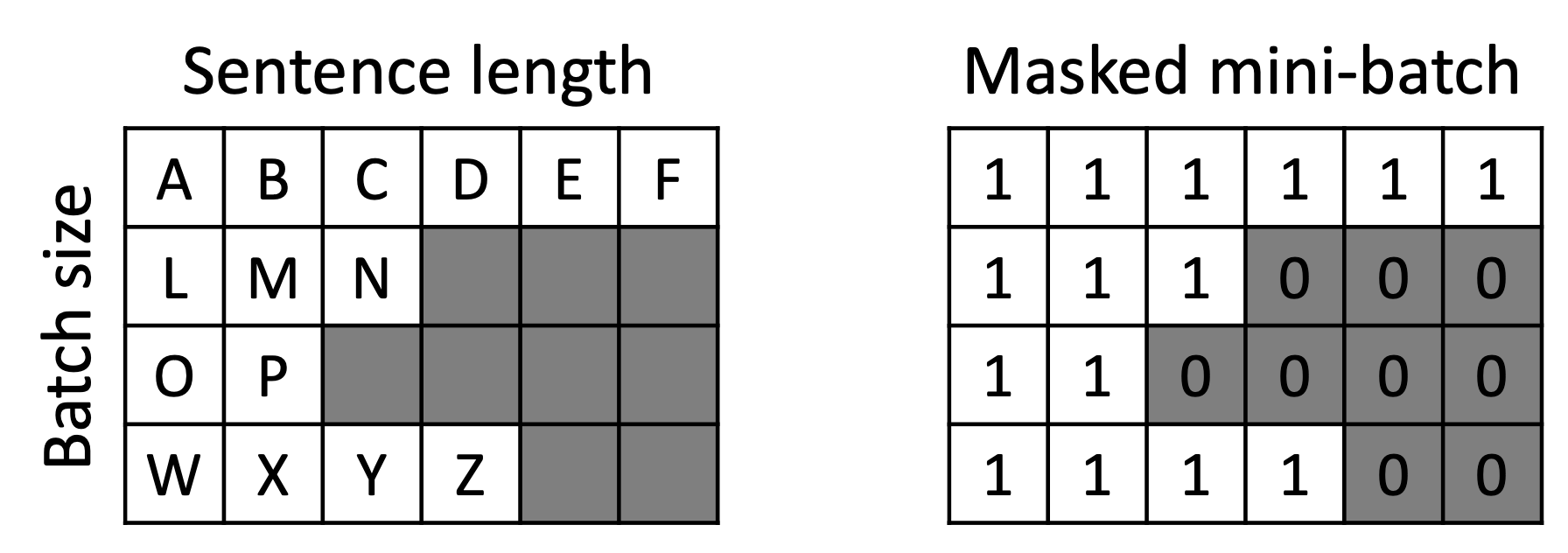}
	\caption{Illustrative example of the inefficiency of the unsorted, randomly sampled mini-batch. In this example, the efficiency of the computation is reduced by $3/8$, since the algorithm spends time computing zeroed tokens (gray) that will be discarded later.}
	\label{fig:mbatch}
\end{figure}

One particular aspect to improve in the Transformer is the computation of empty tokens. Its architecture does not allow the length of the sentences (number of tokens per input) to differ within a mini-batch. The common approach is to define a maximum length within the randomly sampled mini-batch (often the longest sentence's length), and use a padding token to signal the empty tokens of shorter sentences (see Figure \ref{fig:mbatch}). Despite being empty tokens, the Transformer computes their values, and proceeds to discard them afterward. As a result, the model has an extra computational burden that does not contribute to the training of the task.

In this work, we propose an algorithm which sorts translation sentence pairs based on their length before batching. That is, sentences with a similar length will be batched together, thus minimizing the waste of computing power. However, since sorting the whole dataset may break the independent and identically distributed (i.i.d.) assumption of training data samples upon which optimization is formulated, we explore the amount of sorting for mini-batches. That is, instead of sorting the whole data, we sort multiple samples of the mini-batch size, and we can find the trade-off between training time and performance in BLEU score. For example, if we do not sort samples (conventional method), it keeps the i.i.d. assumption while wasting the computation power. With the proposed method, if the mini-batch size is 100 and we sort 1,000 samples while loading the data, then the length of mini-batch makes cycles for every 10 iterations and increases within each cycle, which partially violates the i.i.d. assumption. If we sort 100,000 samples, then there will be cycles for every 1,000 iterations, which breaks the i.i.d. assumption more seriously. 
%%my understanding: If the mini-batch size is 100 and we sort 100,000 samples while loading the data, then the construction of a mini-batch cycles for 1,000 iterations and the length within each cycle increases, which violates the i.i.d. assumption. 

We apply our algorithm to NMT tasks for a high-resource En-Kr (English-Korean), and a low-resource En-Lu (English-Luganda) language pairs. The experimental results demonstrate that using the sorting approach for data loading significantly reduces computation time without sacrificing performance. 

\section{Preliminaries} \label{sec:prel}

\subsection{The i.i.d. assumption in mini-batch training}
In deep learning settings like NLP, training neural networks is an optimization problem that requires finding a set of parameters $\mathbf{\theta}$ such that we minimize the empirical risk $J(\mathbf{\theta}) = \mathbb{E}_{(\mathbf{x}, \mathbf{y})\sim {\hat{p}}} [L(f(\mathbf{x};\theta),\mathbf{y})]$, where $\hat{p}$ is the distribution of the training dataset $\mathcal{D}$, only if $(\mathbf{x}, \mathbf{y})$ are i.i.d., which contributes to the generalization of the trained network \cite{goodfellow2016deep}. 

Instead of processing the whole dataset $\{(\mathbf{x}, \mathbf{y})\}$, $m<<\mathcal{D}$ sample pairs are trained at a time and the total loss in an epoch is the average of the mini-batch loses. This approach is called \textit{mini-batch training}, and it is commonly used in  models such as recurrent neural networks (RNNs) and Transformers to speed up the training process and improve the model's performance. In these settings, it is common to assume that the data within a mini-batch is i.i.d., that is, each sample in the mini-batch is unrelated to the other samples and does not influence the distribution of the other samples, as well as statistics of mini-batches should be unrelated. 

The i.i.d. assumption is important in mini-batch training because it allows the model to make statistical inferences from a random subset of the data, and the learning is representative of the entire dataset. This training strategy is stable, and satisfies the i.i.d. assumption, provided that the mini-batches are sampled randomly. If samples are sorted in any way, during the optimization step, the weights will be updated with some form of bias that degrades the algorithm's generalization ability.

\subsection{Previous works}
Before the era of Transformers, Morishita et al. \cite{morishita2017empirical} conducted a study about the impact of sorting for the Recurrent Neural Network (RNN) NMT algorithm by \cite{luong2015effective}. In their work, the authors studied the impact of \textit{Corpus Sorting Methods}, i.e. sorting the complete input of the model by source/target sentences length before batching. They empirically showed it was faster to train models when sorting, but for the convergence speed and performance it was best to use random shuffle. The authors also showed how the perplexity of the model dropped significantly when sorting compared to random mini-batches. However, the authors hypothesized these results were due to an existent bias in the dataset, in which similar length sentences had very similar content.

In our work, we study the effect of partial sorting the dataset to decrease the effect of non-i.i.d. samples batched together, as explained in the next section. Furthermore, we conducted the study using the Transformer model and NMT datasets that have no content-length correlation.

\section{Methods}

Instead of sorting the whole dataset and thus breaking the i.i.d. assumption, we implement a partial sorting algorithm as follows. We start with a randomly shuffled dataset $\mathcal{D}$. Let $m$ be the size of the mini-batch. We define a new look-ahead hyperparameter $k$, and we read $m\times k$ sentence pairs into a buffer $B$. The algorithm sorts the sentences by the lengths of the source and target, where the source length is the first key and the target length is the second for sorting. Then the data loader returns the top $m$ shortest ones at a time. After reading iteratively, if the buffer has less than $m$ sentence pairs, we read pairs to fill the buffer with $m\times k$ pairs. We repeat the process until the training is finished. For the details of the process, see Algorithm \ref{alg:algorith}. The procedure `iter' is called for every training iteration. 

\begin{algorithm}
\caption{Partial Sorting Algorithm in Data Loader}
\label{alg:algorith}
\begin{algorithmic}
\State \textbf{Class} Data Loader (batch size $m$, look-ahead $k$):
\State \hspace{0.2in} Randomly Shuffled Dataset $\mathcal{D}=\{(\mathbf{x}_j, \mathbf{y}_j)\}$ (where $\mathbf{x}_j$ is a source sentence of length $|x_j|$)
\State \hspace{0.2in} Set an empty buffer $B$ 
\State \hspace{0.2in} \textbf{Procedure} {iter}: 
\State \hspace{0.4in} \textbf{If} the number of elements in $B$ < $m$ \textbf{Then}
\State \hspace{0.6in} Fill the buffer $B$ with $m\times k$ elements $[(\mathbf{x}_i,\mathbf{y}_i)] \in \mathcal{D}$
\State \hspace{0.6in} Sort the elements in $B$ by ($|\mathbf{x}_i|$, $|\mathbf{y}_i|$)
\State \hspace{0.4in} \textbf{EndIf}
\State \hspace{0.4in} Pop the top $m$ elements from $B$ for training 
\State \hspace{0.2in} \textbf{EndProcedure}
\State \textbf{EndClass}
\end{algorithmic}
\end{algorithm}

%\begin{algorithm}
%\caption{Partial Sorting Algorithm in Data Loader}
%\label{alg:algorith}
%\begin{algorithmic}
%\Procedure{Minibatch loading}{}
%    \State Dataset $\mathcal{D}=\{(\mathbf{x}_j, \mathbf{y}_j)\}$ (where $\mathbf{x}_j$ is a source sentence of length $|x_j|$), batch size $m$, look-ahead $k$
%    \State Initialize $buffer = [(\mathbf{x}_i,\mathbf{y}_i)]$ of $m\times k$ elements $\in \mathcal{D}$
%    \For{each $(\mathbf{x}_i, \mathbf{y}_i) \in buffer$}
%        \State Sort each element: $buffer' = [\mathbf{x}'_i,\mathbf{y}'_i]$ such that $|\mathbf{x}'_i|<|\mathbf{x}'_i+1| \forall i \in \{1,...,mk-1\}$ 
%    \EndFor
%    \State Select the top $m$ shortest sentences and return $minibatch = buffer'_{top(m)}$ 
%\EndProcedure
%\State Repeat for whole $\mathcal{D}$
%\end{algorithmic}
%\end{algorithm}

Note that the buffer is sorted according to the length of the source and target sentences $|\mathbf{x}_i|$ and $|\mathbf{y}_i|$. Since it is common that pairwise sentences have similar length, we hypothesize there is no significant difference between sorting by ($|\mathbf{x}_i|$, $|\mathbf{y}_i|$) or ($|\mathbf{y}_i|$, $|\mathbf{x}_i|$).

By partially sorting according to Algorithm \ref{alg:algorith}, we preserve some of the stochasticity of the batching process while still sorting by length to improve the training efficiency. We note that there is a trade-off between the i.i.d. data assumption and computational power with different size of buffer $B$. The extreme cases happen when we randomly shuffle the whole dataset, and then we batch, which results in high computation cost, in opposition to sorting the whole dataset which violates the i.i.d. assumption while speeding the computation time. 
%On the other hand, applying Algorithm \ref{alg:algorith} incurs in a loop, which also takes computational time. The larger the $k$, the more key sorting occurs within the buffer $B$
In practice, we need to determine the value for $k$ (and therefore, $B$) that results in fast(er) computation time whilst maintaining some of the stochasticity of the i.i.d. data. 

% need to talk about trade-off between iid assumption and computational power with different size of buffer. 

\section{Experiments} 
All of our experiments were conducted using the Transformer model as in \cite{vaswani2017attention}. Our algorithm was used for pair-wise translations in two language pairs, one high-resource (English-Korean / En-Kr) and one low-resource (English-Luganda / En-Lu). The experiments were run in the same GPU model (GeForce GTX 1080) for fair comparison.

\begin{figure}[ht]
\centering	
\includegraphics[width=0.65\linewidth]{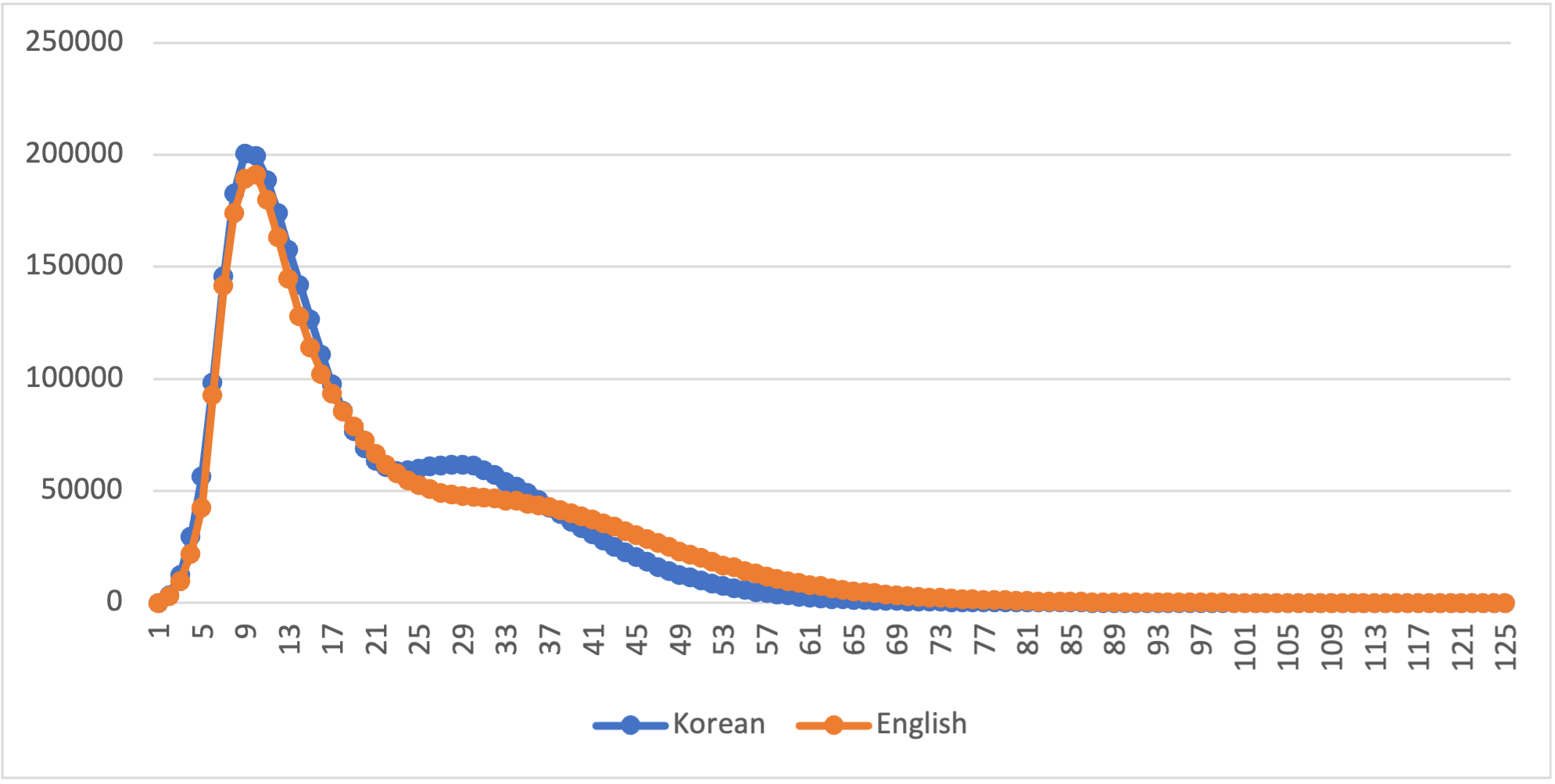}
\caption{Counts of sentences along with the length in the En-Kr dataset. Sentence length was measured by the number of subwords after tokenization, and deleted sentences longer than 125 which were not used in training.}
\label{fig:counts}
\end{figure}

The En-Kr dataset was obtained from the Korpora package via AI Hub\footnote{\url{https://aihub.or.kr/}} and web-crawling\footnote{\url{https://hancorpus.github.io/}}. It consists of $~3.5M$ sentence pairs in various topics, such as spoken language, news, Korean culture, among others. We randomly selected 3,000 pairs for validation and testing, respectively, and used the others for training. 
Table \ref{table:data} presents the average length (number of tokens after BPE tokenization- \cite{sennrich2015neural}) of Korean and English sentences. The table shows English sentences are on average longer than the Korean sentences in terms of number of tokens and that there is a pairwise difference of around two tokens. Since our partial sorting Algorithm \ref{alg:algorith} sorts by both source and target lengths, there is no significant impact of this difference. Fig. \ref{fig:counts} shows the distribution of sentence lengths in the En-Kr dataset. 

For the English-Luganda (En-Lu) language pair, we used the parallel dataset presented in \cite{kimrich2022luganda} which consists of 41,070 sentence pairs. The dataset was split into $96\%/2\%/2\%$ for training, validation and testing, respectively, with a fixed, non-shared vocabulary of 10,000 tokens. As shown in Table \ref{table:data}, after applying BPE tokenization, there is not much difference in the sentence lengths of the languages. 

We conducted a hyperparameter search, and used the best configuration for both language pairs. The best combination was a batch size of 64, a learning rate of $1e-4$ and 256 dimension for the embeddings. The rest of the hyperparameters were left identical to the Transformer architecture presented in \cite{vaswani2017attention}.

\begin{table}[H]
\centering
\begin{tabular}{c|cccc}
\textbf{Language pairs}  & \multicolumn{2}{c}{\textbf{English-Korean (En-Kr)}} & \multicolumn{2}{c}{\textbf{English-Luganda (En-Lu)}} \\ \hline
\textbf{Avg. sentence length}  & \multicolumn{1}{c}{22.64($\pm$15.55)}  & 20.19$(\pm$12.81) & \multicolumn{1}{c}{10.68($\pm$ 3.17)} & 10.67($\pm$3.90)\\ 
\textbf{Pair-wise difference} & \multicolumn{2}{c}{2.45}  & \multicolumn{2}{c}{0.006}  \\ 
\hline                                 
\end{tabular}
\caption{Average sentence length of each parallel corpus after BPE tokenization and average pair-wise length difference between source (English) and target (Korean/Luganda) pairs.} \label{table:data}
\end{table}

\begin{figure}[ht]
\centering	
\includegraphics[width=0.65\linewidth]{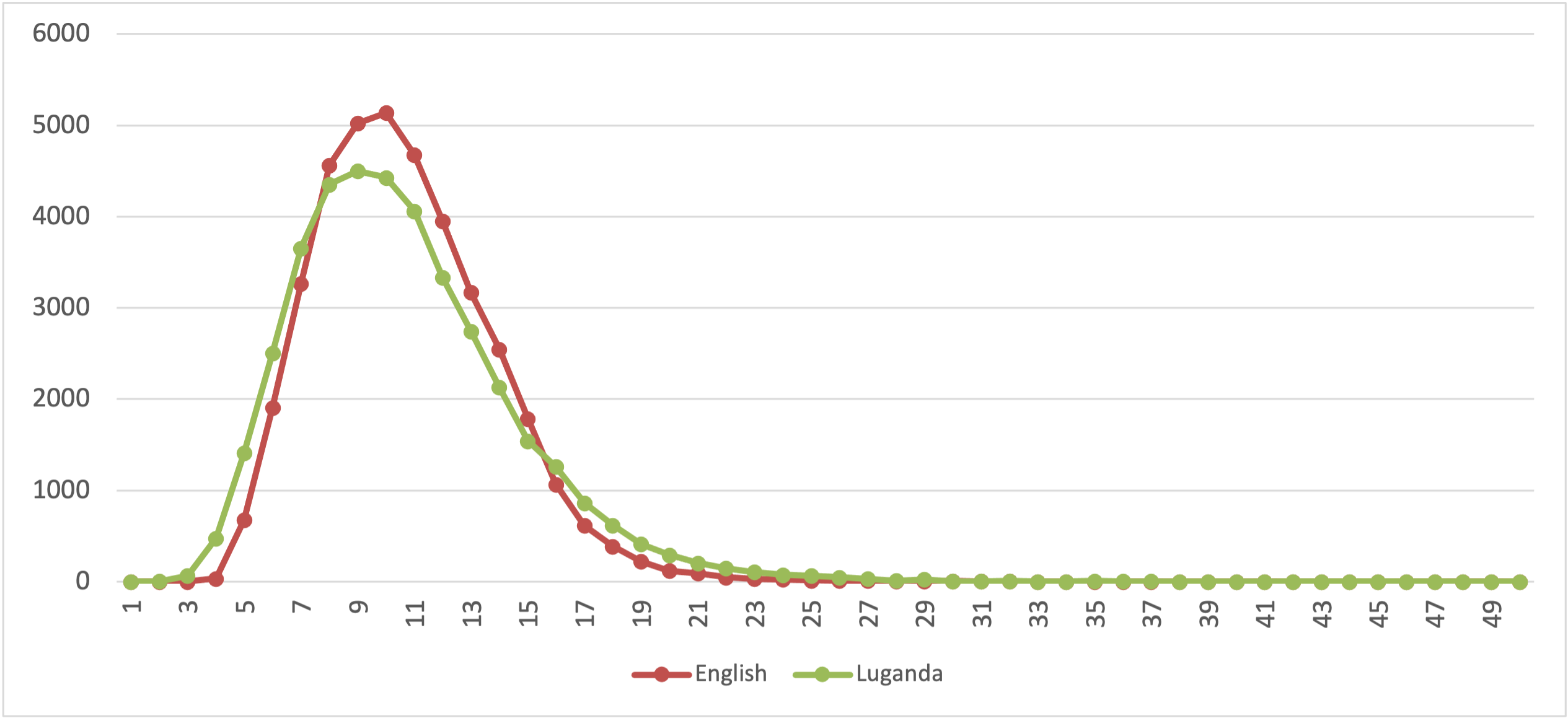}
\caption{Counts of sentences along with the length in the En-Lu dataset. Sentence length was measured by the number of subwords after tokenization, and this dataset \cite{kimrich2022luganda} has sentences no longer than 50 tokens.}
\label{fig:countsenlu}
\end{figure}

As shown in Table \ref{table:data}, on average, the length of the En-Lu pair sentences do not differ significantly. Furthermore, as shown in Fig. \ref{fig:countsenlu}, the sentences lengths are no bigger than 50 tokens after tokenization, and 75\% of the sentences are less than 14 tokens for both languages.

%For both language pairs we performed a hyperparameter search without sorting ($k=1$, i.e. randomly shuffled mini-batches) and chose the best configuration. Next, we performed a hyperparameter search for different $k$ values to study the relation between the proportion of sorted elements in the batch with the overall performance of the model. The results are shown in the next section.

We performed experiments with different $k$ values to study the relation between the proportion of sorted elements in the batch with the overall performance of the model. For both language pairs, all other hyperparameters were fixed for all $k$ values. The results are shown in the next section.

%\section{Results}

%To check the efficiency of the proposed method, we trained neural machine translation models with the En-Kr and En-Lu datasets, and report the training time and translation performance in BLEU scores. 

\subsection{Results with the En-Kr dataset}

We experimented with three look ahead $k$ values; $k=\{1, 1000, 2500\}$ (where $k=1$ is unsorted/random shuffle training). Sorting with $k=1,000$ means that at every 1,000th iteration, Algorithm \ref{alg:algorith} sorts 1.8\% of the overall dataset, and every iteration it returns the top 64 shortest sentences remaining in the buffer. For $k=2,500$, the algorithm sorts 4.5\% of the data at a time. 
%Sorting with $k=1,000$ means that at each iteration, Algorithm \ref{alg:algorith} is sorting 4\% of the overall dataset before selecting the top 64 shortest sentences. For $k=2500$, the algorithm sorts 10\% of the data at a time. 

The results for the En-Kr dataset are shown in Table \ref{table:enkr_bleu}. Overall, the performances in terms of the BLEU score are quite similar, although there is a slight negative correlation with the $k$ hyperparameter. However, in terms of efficiency, the unsorted (conventional) cases take twice as long compared to $k=1,000$ (ahead) cases, and $k=1,000$ and $k=2,500$ cases take roughly the same time. 

\begin{table}[H]
\centering
\begin{tabular}{clccc}
\multicolumn{2}{c}{\textbf{Ahead ($k$)}}  & \textbf{unsort} & \textbf{1000}                      & \textbf{2500}  \\ \hline
\multirow{3}{*}{\textbf{En-Kr}} & \textbf{Avg. time {[}ms{]}} & 224.28           & 117.34                             & 112.82        \\
                            & \textbf{BLEU valid.} & 19.57 & 19.55  & 19.40 \\ 
                            & \textbf{BLEU test} & 19.27 & 19.39 & 19.29 \\ \hline
\multirow{3}{*}{\textbf{Kr-En}} & \textbf{Avg. time {[}ms{]}} & 220.44           & 115.88                             & 112.80      \\
                            & \textbf{BLEU valid.} & 39.08  & 38.87 & 38.46 \\
                            & \textbf{BLEU test} & 37.38  & 37.79 & 37.73 \\
\hline                                  
\end{tabular}
\caption{BLEU scores of the En-Kr pairs with different $k$ from no sorting ($k=1$) to sorting with $k=\{1000,2500\}$. We show the average time (in $ms$) for each iteration.} \label{table:enkr_bleu}
\end{table}

%From the performance, we can observe the expected trade-off between the generalization capability and the computational cost of the training. Partially sorting the mini-batch incurs in a significant decrease of computational time, but also adds a marginal degradation of the BLEU score. The more data samples we sort at each iteration (bigger $k$), the more we harm the performance.

Furthermore, as shown in Table \ref{table:avg_batch_len}, the case of $k=1,000$ has a decrease of almost 65\% average sentence length per mini-batch than the unsorted case, and similar to the $k=2,500$ case. Between $k=\{1000,2500\}$, the higher $k$ has a relative degradation of BLEU score of 0.6\%, whilst having  
% a 3.8\% 
a marginal gain in computational cost. Therefore, in this study case we would empirically choose $k=1,000.$

\begin{table}[H]
\centering
\begin{tabular}{c|cc}
\textbf{Ahead} ($k$) & \textbf{Avg. length in Korean} & \textbf{Avg. length in English}\\ \hline
\textbf{1 (unsorted)} & 61.37 & 72.34 \\
\textbf{1,000} & 22.25 & 26.94 \\
\textbf{2,500} & 22.21 & 25.85 \\
\hline                                 
\end{tabular}
\caption{Average batch lengths with different sorting sizes for Kr2En translation. By sorting with $k=1,000$, the length of minibatch becomes 36\% of the `unsorted' case, which saves a lot of computation power. Even when $k$ increases further into 2,500, however, the batch length does not significantly decrease.} 
\label{table:avg_batch_len}
\end{table}

\subsection{Results with the En-Lu dataset}
For the En-Lu pairs, we experimented with more $k$ values, taking into advantage that the dataset size does not require as many training days as the En-Kr pair. We chose $k=\{1,100,250,500, all\}$, which account to no sorting, partial sorting $15\%$, $39\%$, $78\%$ of the dataset and finally, sorting the whole dataset before mini-batching ($all$.) We show in Table \ref{table:enlu_bleu} that, similarly to the En-Kr case, the BLEU score does experience a marginal performance boost in most cases. However, the relative decrease of computation time for each mini-batch is lower than the En-Kr case. We believe that the maximum sentence length (50) is too small to enjoy significant improvement by minimizing the waste of computation power for the padding tokens.  
\begin{table}[H]
\centering
\begin{tabular}{clccccc}
\multicolumn{2}{c}{\textbf{Ahead ($k$)}} & \textbf{unsorted} & \textbf{100} & \textbf{250} & \textbf{500} & \textbf{all data} \\ \hline
\multirow{3}{*}{\textbf{En-Lu}} & \textbf{Avg. train time {[}ms{]}} & 57.82 & 53.17  & 53.27  & 52.79 &  50.73 \\
                                & \textbf{BLEU valid.}  & 18.69 & 19.14 & 18.94   & 18.58  &  19.53 \\
                                & \textbf{BLEU test}  & 16.91 & 17.55 &  17.04   & 17.34  &  17.22      \\
                                 \hline
\multirow{3}{*}{\textbf{Lu-En}} & \textbf{Avg. train time {[}ms{]}} & 56.08 & 53.88  & 55.25 & 47.74  & 47.34 \\
                                & \textbf{BLEU valid.}  & 23.82 & 23.41 & 23.60 &  23.07   & 23.55 \\
                                & \textbf{BLEU test}  & 22.34 & 21.02 & 23.62  &  22.65   & 23.23        \\     
\hline                                 
\end{tabular}
\caption{BLEU scores of the En-Lu pairs with different $k$ from no sorting ($k=1$) to sorting the whole dataset (`all data'.) We show the average time (in $ms$) it took the model to process each mini-batch.} \label{table:enlu_bleu}
\end{table}

As seen in Table \ref{table:avg_batch_lenenlu}, the average length within a batch is down to around 40-55\%. This is in correlation to the fact that, as showed in Figure \ref{fig:countsenlu}, most of the sentences have similar length, so the gain is not as significant as a larger dataset like the En-Kr case. Regardless, there is still an improvement in the efficiency of the computation and task performance.

\begin{table}[H]
\centering
\begin{tabular}{c|cc}
\textbf{Ahead ($k$)}       & \textbf{Avg. length in Luganda} & \textbf{Avg. length in English} \\ \hline
\textbf{1 (unsorted)} & 25.30                           & 22.23                           \\
\textbf{100}             & 14.71                           & 12.93                           \\
\textbf{250}             & 13.28                           & 12.30                           \\
\textbf{500}             & 12.63                           & 12.10                          \\
\textbf{all}             & 11.54                           & 11.28                           \\ \hline
\end{tabular}
\caption{Average batch lengths with different sorting sizes for Lu2En translation. }%The largest relative decrease in length occurs in every other $k$ with the unsorted case ($k=1$).} 
\label{table:avg_batch_lenenlu}
\end{table}
    
\section{Conclusion}
In this work, we explored the effect of partial sorting of dataset by sentence length during the training of translation models. We introduced a look-ahead hyperparameter $k$ that reads $k$ times the batch size into a buffer, after which it would sort and return the top $k$ shortest sentences. 
The experiment results show that the proposed partial sorting approach significantly reduces computational time for processing mini-batches in En-Lu and En-Kr translation tasks, while keeping the same or even slightly better performance in the BLEU score. Especially, for a relatively large En-Kr dataset, the processing time becomes around half of the original random shuffled case. 
The presented method is an effective and easy approach that can be integrated to any training process with flexible data length and will have a noticeable effect in the training efficiency. 

%In this work, we explored the effect of partial sorting of dataset by sentence length during the training of translation models. We introduced a look-ahead hyperparameter $k$ that read $k$ times the batch size into a buffer, after which it would sort and batch the top $k$ shortest sentences. This way, we could empirically control the trade-off between randomly shuffling and batching (which preserves the i.i.d. assumption at the expense of higher computational cost), and completely sorting the dataset (which trades efficiency with the violation of the i.i.d. assumption.) 
%Our work shows that for a relatively large, diverse dataset such as the Korpora package Kr-En dataset, there are gains of almost half of the computational time per iteration than just randomly shuffling. Furthermore, the performance is not harmed, and in some cases it even increases the BLEU scores. For a low-resource, smaller data like \cite{kimrich2022luganda}, we also empirically showed favorable results, despite the dataset not being diverse enough w.r.t. sentence lengths. The presented partially sorting method is an effective and easy approach that can be integrated to any Transformer architecture and will have a noticeable effect in the training efficiency. We hope that the research community will consider this as an alternative to the existing training methods.

\section{Acknowledgments}
This research was supported by Basic Science Research Program through the National Research Foundation of Korea funded by the Ministry of Education (NRF-2022R1A2C1012633), and by Institute for Information \& communications Technology Promotion (IITP) grant funded by the Korea government(MSIT) (No. 2018-0-00749, Development of virtual network management technology based on artificial intelligence).

\bibliographystyle{abbrv}
\bibliography{refs}

\end{document}